\documentclass[pdflatex,sn-basic,iicol]{sn-jnl}

\usepackage{graphicx}
\usepackage{booktabs}
\usepackage{multirow}
\usepackage{array}
\usepackage{amsmath,amssymb}
\usepackage{microtype}
\usepackage{xcolor}
\usepackage{url}
\usepackage{dblfloatfix}
\usepackage{placeins}
\newenvironment{revision}{\begingroup\color{red}}{\endgroup}

\newcommand{\rev}[1]{#1}
\renewenvironment{revision}{}{}

\begin{document}

\title[DiVA: Interactive Digital Life Simulation]{DiVA: Enabling Interactive Digital Life Simulation via Video Models}

\author[1,2,3]{\fnm{Cheng} \sur{Chen}}
\author[3]{\fnm{Hao} \sur{Ouyang}}
\author[3]{\fnm{Qiuyu} \sur{Wang}}
\author[3]{\fnm{Ka Leong} \sur{Cheng}}
\author[3]{\fnm{Wen} \sur{Wang}}
\author[3]{\fnm{Yihao} \sur{Meng}}
\author[3]{\fnm{Hanlin} \sur{Wang}}
\author[3]{\fnm{Yixuan} \sur{Li}}
\author[1]{\fnm{Jiacheng} \sur{Wei}}
\author[4]{\fnm{Zhenshan} \sur{Tan}}
\author[3]{\fnm{Yanhong} \sur{Zeng}}
\author[3]{\fnm{Yujun} \sur{Shen}}
\author[1]{\fnm{Guosheng} \sur{Lin}}
\author*[2]{\fnm{Fayao} \sur{Liu}}
\email{liu\_fayao@a-star.edu.sg}

\affil[1]{
  \orgname{Nanyang Technological University},
  \orgaddress{\country{Singapore}}%
}

\affil[2]{
  \orgname{Institute for Infocomm Research, A*STAR},
  \orgaddress{\country{Singapore}}%
}

\affil[3]{
  \orgname{Ant Group},
  \orgaddress{\country{China}}%
}

\affil[4]{
  \orgname{Nanjing University of Information Science and Technology},
  \orgaddress{\city{Nanjing}, \country{China}}%
}

\abstract{We present DiVA, a deeply interactive digital life simulator pioneering a new paradigm for long-term, open-ended interactive experiences within digital character worlds. DiVA's architecture pairs a Multimodal Large Language Model (MLLM) as a router with a meticulously designed stacked video pipeline for seamless, multi-turn interactions with action and audio response. To maintain continuity and avoid degradation, we model generation as a three-part coupled system: waiting video, action video, and the transitions between them. These transitions are critically handled by our Anchored Video Continuation (AVC) module, which returns the character to stable states to prevent degradation. By encoding information from the preceding action video segment, AVC ensures smooth transitions, significantly reducing camera jitter and inconsistencies common in current video transition methods. This design also enables complex pose changes (e.g., sitting to standing) typically difficult for audio-driven models. These system designs together ensure high-fidelity identity, coherence, and dynamics for extended experiences.  To validate our pipeline design, we comprehensively compare our system against alternatives by replacing our core generation module with mainstream long-video, continuation, and interpolation methods. \rev{We further analyze the necessity of the three-stage design, anchor-state selection, transition naturalness, spatial grounding, and the quality--latency trade-off, and we expand the comparison to additional long-form audio-driven avatar models.} Results confirm DiVA is markedly superior in maintaining long-term visual quality and realism, validating its effectiveness as a sustainable, interactive simulation.}


\maketitle

\begin{figure*}[!t]
    \centering
    \includegraphics[width=0.96\textwidth]{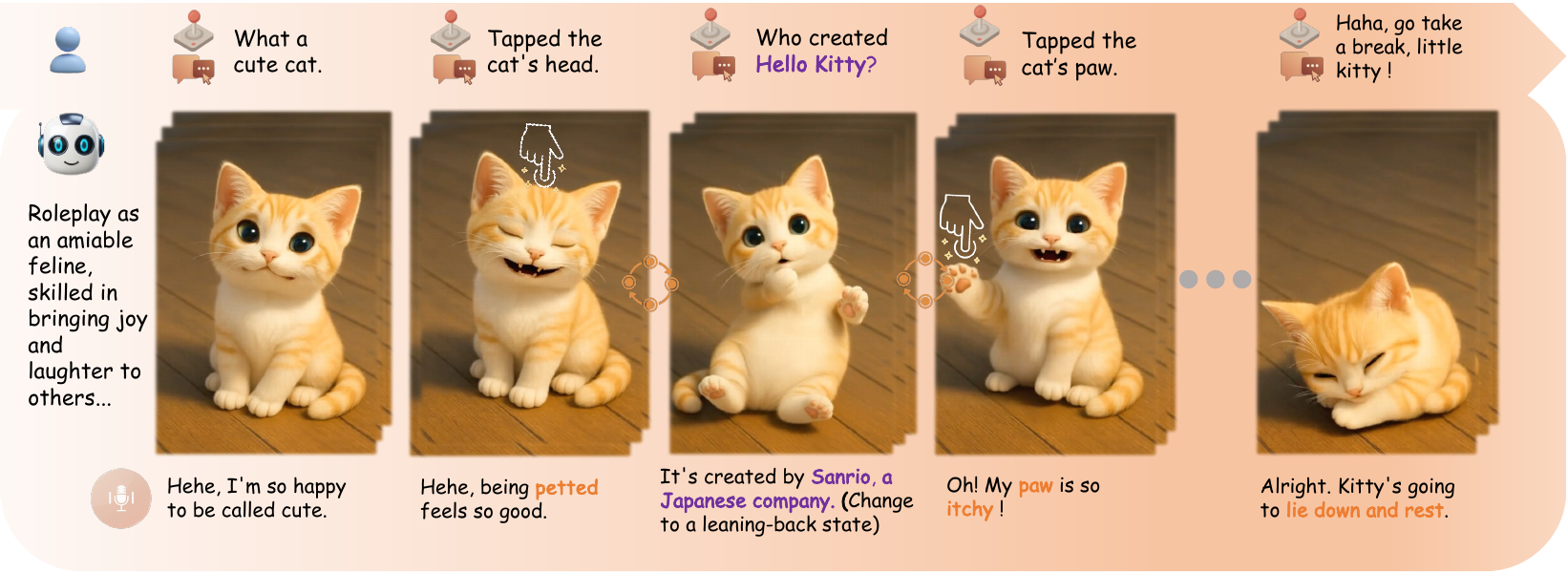}
    \caption{Our DiVA enables infinite, continuous interaction with undiminished quality (five rounds shown here). Users can define character personality and interact via simultaneous verbal and non-verbal (spatial click) inputs, receiving corresponding realistic actions and voice replies. DiVA demonstrates advanced state transition capabilities (3rd interaction), while also inheriting the MLLM's strong ability for factual responses (e.g., describing Hello Kitty), enabling limitless simulation within interactive digital worlds.}
    \label{fig:teaser}
\end{figure*}

\section{Introduction}
\label{sec:intro}

Pushing video generation\cite{wan2025wan, kong2025let, bai2025recammaster}, from creating short and passive clips to simulating long-term and interactive digital lives, represents an exciting new frontier for the field. We frame this vision as Digital Life Simulation (DLS): all behaviors, visuals, and dialogue are generated on demand by models, with no predefined rules or handcrafted assets, as shown in Figure~\ref{fig:teaser}. Users interact through open-ended language and spatial clicks, while the system maintains long-term persona and memory and continuously produces role-centric response shots aligned with user intent.

This idea resonates with recent discussions of infinite games, where experiences evolve without preset boundaries or fixed rules. This concept has been explored by generating image series based on evolving prompts \cite{li2024unbounded}. Taking this a step further, previous work \cite{cheng2025animegamer} lets users interact with characters in a virtual world through language and generates multi-turn states composed of dynamic animation shots, moving toward openness. However, it mainly adopts a third-person perspective of the character and produces isolated, non-contiguous video segments, which makes it hard for users to embody the character and limits conversational, continuous video expression and coherence.

To address these gaps, we present DiVA: a deeply interactive digital life simulator that maintains high dynamics and strong expressivity over long-form, multi-turn interactions while closely tracking user input intent. Specifically, we designed both linguistic (language) and spatial (e.g., spatial clicks) interactions to allow users to engage more immersively with the character. During long-term interaction, the system maintains the character’s high-fidelity visual identity and long-range persona, and continuously renders role-centric response shots. These shots synchronize the character's spoken dialogue with non-verbal actions (e.g., gestures, expressions) that align with the user's interaction input.

The recent advent of Multimodal Large Models (MLLMs) \cite{bai2023qwen, achiam2023gpt} has brought significant advancements to role-playing, achieving high levels of personalization and coherence using in-context learning \cite{dong2022survey}. Motivated by these advancements, the MLLM in DiVA functions as the simulation router. Initially, it is conditioned via in-context exemplars to instantiate the character’s persona and interaction environment. Subsequently, during the multi-turn interaction, it reasons over user input (both linguistic and spatial) and the interaction history to produce structured directives. These directives jointly encode the character’s dialog response and the behavioral prompts for the video generator, directing it to render the precise, synchronized response shots mentioned earlier.

Relying on a single, monolithic video generator presents a significant gap in achieving both high-quality expressivity and long-term stability, while also severely limiting controllability. To address these limitations and achieve naturalness, expressivity, and long-term stability, we design a three-part coupled video generation pipeline. First, waiting video produces subtle idle motion at the start to avoid a frozen avatar feel. This is followed by Action Video, which performs MLLM-guided audio–video alignment to generate gestures and facial expressions tightly coupled to speech.
\begin{table*}[!t]
\centering
\caption{\rev{Positioning of representative video generation paradigms for DLS, including E2E general long video generation 
(FramePack, LongLive, LongCat-Video), E2E Audio-Avatars 
(InfiniteTalk, StableAvatar, WanS2V, Live Avatar, X-Streamer)}}
\label{tab:paradigm}
\scriptsize
\setlength{\tabcolsep}{3pt}
{%\color{red}
\begin{tabular*}{0.96\textwidth}{
    @{\extracolsep{\fill}}lcccc@{}
}
\toprule
\textbf{Paradigm}
& \textbf{\shortstack{Speech\\Alignment}}
& \textbf{\shortstack{Macro\\Transition}}
& \textbf{\shortstack{State\\Routing}}
& \textbf{\shortstack{Quality\\Reset}} \\
\midrule
E2E General long video generation
& N/A
& Unstable
& None
& None \\

E2E Audio-avatars
& Strong
& Restricted
& Implicit
& None \\

\textbf{Decoupled Stateful Pipeline}
& \textbf{Strong}
& \textbf{Robust}
& \textbf{Explicit}
& \textbf{Anchored} \\
\bottomrule
\end{tabular*}}
\end{table*}
To handle the critical transitions between these multi-turn interactions, we introduce anchored video continuation (AVC). We found that simple interpolation from the prior video's last frame causes camera and motion discontinuity. Our AVC therefore instead conditions on a segment of the preceding action video while targeting on preset, high-quality anchor frames. This mechanism allows the character to reset to a high-quality state after each action, which is essential for suppressing identity drift. Critically, the AVC phase is decoupled from strict audio alignment, which limits its expressive potential. This allows our method to expand the character's motion range for more creative dynamic effects, such as performing natural pose changes concurrently with camera movement that is difficult to achieve in the tightly-coupled action video phase.

To validate our pipeline design, we conduct a comprehensive comparison. We benchmark our system against alternative approaches by systematically substituting our core video module with mainstream long-video generation, state-of-the-art continuation, and interpolation-based methods. These results demonstrate that in long-duration, multi-turn digital simulations, DiVA is significantly superior in maintaining visual quality, character response accuracy, and high-dynamics behavior. These capabilities validate our design's effectiveness, providing a robust foundation that realizes an infinite exploration and pushes generative media toward a future of persistent, interactive digital worlds.

\section{Related Work}
\label{sec:related}

\subsection{Video Generation Models}
\label{sec:rw_video}

Inspired by the scaling laws observed in language models, a surge of works \citep{polyak2024movie,kong2024hunyuanvideo,blattmann2023stable,wan2025wan,gao2025seedance,OpenAI_Sora2_2025,Google_Veo3_2025,Kuaishou_Kling_2025} has begun to extend the capabilities of video generative models through scalable data curation strategies and large-scale model training. Benefiting from the extensive training scale, these models exhibit impressive visual quality.

\begin{revision}
Recent systems further improve the temporal prior through larger diffusion-transformer backbones, stronger visual--text encoders, and multi-stage training. These foundation models provide the high-fidelity appearance and motion prior which are primarily optimized for bounded clips. Repeatedly applying a short-video generator does not by itself create a persistent simulation, because each imperfectly generated segment becomes the condition for subsequent turns and local errors can therefore accumulate.
\end{revision}

Long-video generation is critical for applications requiring long-term coherence, such as embodied AI and world models. Key strategies to address generation challenges include enhancing robustness by injecting and denoising noise (e.g., Rolling Diffusion; compressing historical frames into a fixed latent \citep{zhang2025test,cai2025mixture} or combining compression with future frame planning (e.g., FramePack \citep{zhang2025packing}); and using task-specific pretraining to prevent color drift (e.g., LongCat-Video \citep{team2025longcat}). Additionally, domain-specific models exist, such as InfiniteTalk \citep{yang2025infinitetalk}, which specializes in generating long talking-body videos. In our experiments, we compare our pipeline against the aforementioned approaches, demonstrating its effectiveness during ultra-long, turn-by-turn video generation.

\begin{revision}
LongLive adopts causal frame-level autoregression with KV-recache and streaming long tuning to support real-time prompt changes \citep{yang2025longlive}, whereas HoloCine models a complete multi-shot scene through shot-localized text control and sparse inter-shot attention \citep{meng2026holocine}. These methods substantially extend duration and cross-shot coherence, but most still follow a predominantly single-stream generation trajectory: the current generated state is repeatedly reused as future context, and restoration of a clean visual state is handled only indirectly through prompts, memory, or context compression. iMontage provides a complementary example in which the temporal prior of a pretrained video backbone is repurposed for variable-cardinality many-to-many image generation \citep{fu2026imontage}. This supports a broader principle also adopted by DiVA: strong pretrained video priors can be reorganized through task-specific structure rather than replaced by a monolithic model trained from scratch.
\end{revision}

\subsection{Controllable Video Synthesis}
\label{sec:rw_control}

Concurrently, a parallel line of research has focused on enhancing the controllability of video generation through various modalities. For instance, some works \citep{geng2025motion,zhang2025tora,yin2023dragnuwa} utilize trajectories to guide the generation process, while others \citep{bai2025recammaster,luo2025camclonemaster,bahmani2025ac3d} concentrate on the control of cameras. Although these methods enable precise control, it is typically limited to single, short-form video clips and they are not designed for immersive, long-duration interaction.

\begin{revision}
Recent controllable systems \citep{chen2025cadcrafter,chen2024sculpt3d} broaden the conditioning interface. WorldCanvas jointly uses reference images, trajectories, and text to specify the identity, timing, visibility, and semantics of promptable world events \citep{wang2025worldcanvas}; AnyWorld factorizes egocentric interaction into action, camera, and embodiment controls to recompose human experience into robot-native rollouts across embodiments and viewpoints \citep{chen2026anyworld}; Human-animation methods additionally condition on audio, poses, end effectors, or reference identities to obtain semantically meaningful character motion \citep{tian2025emo2,kong2025let,gao2025wan,meng2025echomimicv2}. These studies demonstrate that additional conditions can substantially improve local control, but sustained interaction further requires the system to decide when the visual state should change and how quality should be restored after many turns.

First--last-frame generation and video continuation provide temporal endpoint control closely related to our transition problem. A target frame specifies the desired terminal appearance, while preceding content constrains short-term continuity. Standard first--last-frame interpolation, however, normally represents the starting state with only one frame. In an interactive setting, the terminal frame alone is an incomplete description of ongoing motion: two clips may end at visually similar frames while having different body momentum, gesture phase, and camera movement. DiVA therefore conditions AVC on a preceding video segment, so that the generated transition behaves as a continuation of the ongoing motion rather than a newly initialized interpolation.
\end{revision}

\subsection{Digital Life Generation}
\label{sec:rw_dls}

\noindent\textbf{Talking Avatar Generation.}
Many pioneering works have focused on endowing AI with a visual embodiment. As speech is a critical medium for character simulation, research has predominantly centered on generating talking faces or talking bodies. The generation of talking avatars can be broadly categorized into two approaches: 3D-based \citep{chan2022efficient,qiu2025lhm,qian2024gaussianavatars} and video-based \citep{kong2025let,tian2025emo2,gao2025wan,meng2025echomimicv2}. 3D-based methods typically rely on an intermediate 3D representation for animation, a process that often results in lower visual fidelity and naturalism. In contrast, video-based methods leverage end-to-end training with large-scale data curation, which tends to yield superior realism in facial expressions and body dynamics. Despite their strong performance in lip-syncing scenarios, generating avatars with a consistent personality and natural long-term motion still requires an additional process.

\begin{revision}
Long-form audio-driven systems such as InfiniteTalk, StableAvatar, WanS2V, and Live Avatar extend this paradigm through sparse reference frames, sliding windows, causal adaptation, or streaming inference \citep{yang2025infinitetalk,tu2025stableavatar,gao2025wan,huang2025liveavatar}. X-Streamer further integrates multimodal reasoning and audiovisual generation in a Thinker--Actor architecture for persistent video-call interaction \citep{xie2025xstreamer}. These systems are highly relevant to conversational avatars, but their primary objective remains continuous audio-conditioned rendering. Strong audio coupling benefits lip synchronization and local conversational motion, whereas digital life simulation additionally requires deliberate macroscopic state changes, spatial interaction, and explicit long-horizon quality control.
\end{revision}

\noindent\textbf{AI Agents in Games and Simulations.}
There has been a line of exploration in enabling AI to simulate various roles in games, such as competitors, designers, or teammates \citep{zhu2021player,pell1992metagame}. For instance, Unbounded \citep{li2024unbounded} generates an infinite game world, utilizing an LLM to generate text responses and a pre-trained text-to-image model enhanced with LoRA \citep{hu2022lora} for character-consistent image generation. Building on this, AnimeGamer \citep{cheng2025animegamer} extends the concept to anime life simulation, employing a video model to generate dynamic animation shots. While both Unbounded and AnimeGamer utilize a Multimodal Large Language Model (MLLM) \citep{achiam2023gpt,bai2023qwen,liu2024deepseek} as a router to transform dialogue into character information, their limitations distinguish them from our work. Unbounded is primarily focused on static image generation. AnimeGamer, despite using a video model to simulate character movements, mainly adopts a third-person perspective and produces isolated video segments. This approach makes it hard for users to embody the character and limits interaction-driven, continuous video expression.

\begin{revision}
Multi-turn visual interaction has also been explored through visual dialog and interactive character systems \citep{chen2022utc}. Taken together, existing studies establish language models as effective interaction agents, but leave an important rendering-side question: how can open-ended multimodal intent be converted into continuous, speech-aligned video while retaining both expressive large motion and stable visual identity over many turns?

We view existing long-form character generation as two complementary paradigms. General long-video models provide broad motion, but do not explicitly route interaction states or deliberately restore a vetted visual state. Audio-driven avatar models provide strong speech alignment and local identity preservation, but large pose and camera transitions remain constrained by the audio-coupled generation objective. DiVA addresses this stability--expressiveness trade-off with a decoupled stateful pipeline: action video handles speech-aligned responses, AVC performs audio-decoupled macroscopic transitions, and curated anchors provide explicit visual reset states selected by the MLLM. Table~\ref{tab:paradigm} summarizes this positioning.
\end{revision}

\section{Methods}
\label{sec:method}

\subsection{Task Formulation}

\begin{figure*}[!t]
    \centering
    \includegraphics[width=0.96\textwidth]{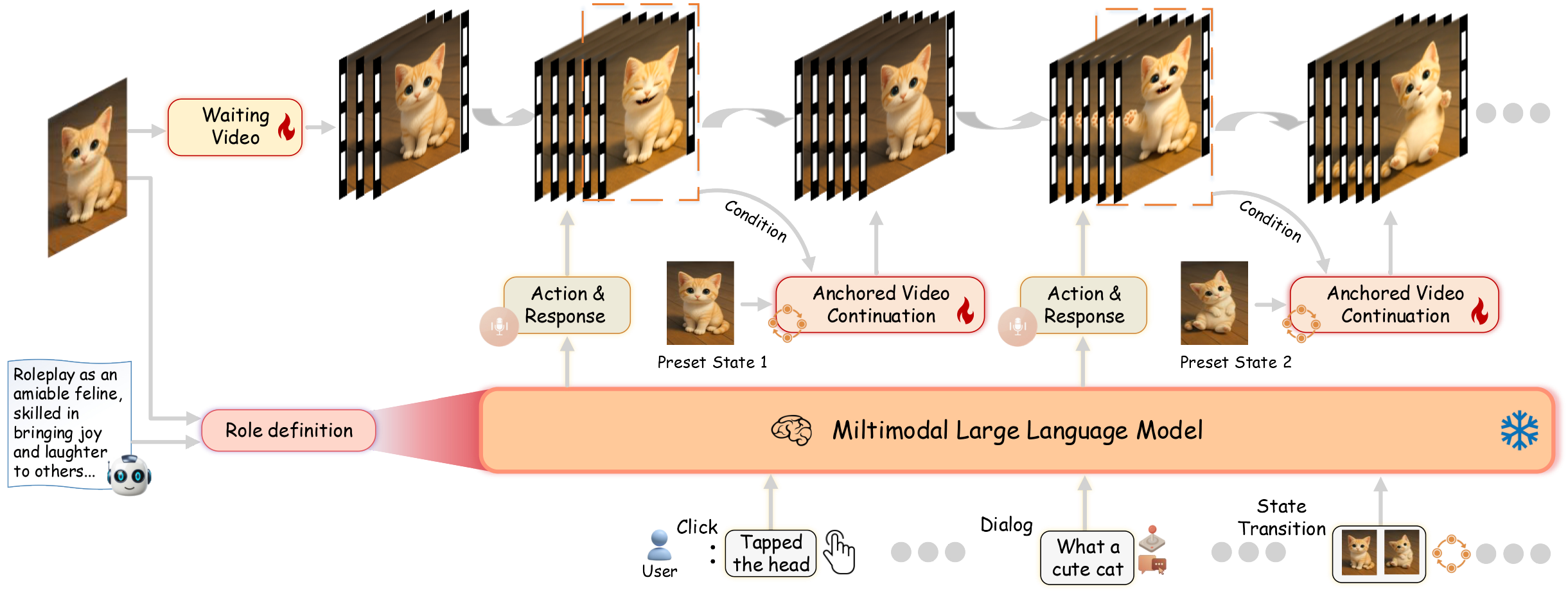}
    \caption{Given a text prompt defining the character's role, an MLLM acts as a router, processing user inputs (e.g., clicks and dialogue) and translating them into character responses and behavioral prompts for the audio-text conditional action model. Our video generation component starts with a waiting video, followed by iterative action videos based on the MLLM output. Finally, the anchored video continuation (AVC) module, conditioned on the prior action, targets preset, high-quality anchor frames that represent distinct character states (e.g., sitting or leaning back). Guided by the MLLM state assessment, AVC transitions to the appropriate anchor state. This mechanism smoothly restores the sequence to a stable, high-quality condition and enables expressive transitions between states, significantly expanding the character's motion range.}
    \label{fig:overview}
\end{figure*}

As shown in Figure~\ref{fig:overview}, we focus on the challenging task of digital life simulation in this paper, wherein a digital character is expected to deliver an enduring, consistent, and dynamic immersive interactive simulation guided by both a user-defined personality and ongoing user input. We employ a Multimodal Large Language Model (MLLM) as a router to process user input and translate it into a corresponding character response for the generative models. Subsequently, a video model is utilized to render the actions of the digital character. To support natural, long-term, and multi-turn action generation, our video generation model is composed of three key modules: a waiting video module, an action video module, and an anchored video continuation module.

\begin{revision}
To make the stateful interaction process explicit, let $\mathcal{C}$ denote the static context, including the character identity, persona, scene, voice configuration, and a portfolio of anchor states $\mathcal{A}=\{I^a_k\}_{k=1}^{K}$. At turn $t$, the user provides multimodal input $U_t$, which may contain language and a spatial click, while $H_{t-1}$ stores the interaction history. In addition to the generated video, DiVA maintains a semantic state selected from $\mathcal{A}$. This differs from ordinary long-video continuation, where the system state is represented only by the most recently generated frames and therefore inherits their accumulated errors.
\end{revision}

\subsection{MLLM as Digital Life Simulation Router}

The recent advent of MLLMs has brought significant advancements to role-playing, achieving high levels of persona fidelity and temporal coherence. Our empirical evaluations revealed that MLLMs possess potent zero-shot and few-shot capabilities, where an In-Context Learning (ICL) \citep{dong2022survey} strategy achieves strong performance without costly fine-tuning. Therefore, for architectural parsimony and generalization, we adopt the ICL approach in our simulation. In our implementation, we primarily leverage large models such as GPT-5 \citep{achiam2023gpt} and Qwen \citep{bai2023qwen}. We observed that robust performance is achievable with a concise set of ICL exemplars.

In our framework, the MLLM functions as the central semantic orchestrator. Its responsibilities begin prior to the simulation, where it is conditioned with a series of descriptive exemplars defining the interaction context, the character's emotional range, and its dialogic style. Our framework also supports visual persona definition. Users can provide a single reference image, which is then processed by a generative vision module (Gemini \citep{gemini2024gemini} or Qwen \citep{bai2023qwen}) to synthesize a portfolio of anchor states---visual representations of the character in distinct emotional or physical states using instruction-based editing.

Subsequently, the MLLM manages the turn-by-turn dynamics. Critically, it performs unified reasoning over multimodal user input (linguistic and spatial) and the interaction history. This process yields the character's response, including the dynamic selection of the appropriate anchor state from the pre-computed visual state-space, which triggers coherent visual transitions. The MLLM output at each turn is a structured semantic output $O_t=(P^B_t,R^D_t)$, jointly encoding the behavioral prompt $P^B_t$ and the dialogic response text $R^D_t$.

This multi-turn process, governed by the MLLM $\mathcal{M}$, can be formalized as the following auto-regressive function at each turn $t$:
\begin{equation}
O_t = \mathcal{M}(U_t,H_{t-1},\mathcal{C}),
\label{eq:mllm}
\end{equation}
where $\mathcal{C}$ is the static context, $U_t$ is the current multimodal user input, $H_{t-1}$ is the interaction history, and $O_t=(P^B_t,R^D_t)$ is the structured output.

For the spatial instruction modality, we explored two grounding mechanisms. A naive approach of providing raw click coordinates proved ineffective, as it requires the MLLM to solve a difficult numerical-to-visual grounding problem. Our second, more robust approach transforms the spatial coordinate task into a visual-referential grounding problem. When a user interaction is detected, we dynamically augment the current video frame with a salient visual marker (e.g., a dilated red dot) at the point of interaction. This augmented frame is then fed to the MLLM. Simultaneously, the ICL prompt is updated to condition the model to interpret this visual marker as the locus of user intent (e.g., the user is clicking the red dot). This method obviates the need for explicit coordinate mapping, instead leveraging the MLLM's potent visual-contextual reasoning capabilities to infer the user's intent.

\begin{revision}
More formally, given the current frame $I_t$ and a click coordinate $(x_t,y_t)$, we construct
\begin{equation}
\widetilde{I}_t=\operatorname{DrawMarker}(I_t;x_t,y_t,r),
\end{equation}
where $r$ is the marker radius. The MLLM then performs ordinary visual referring on $\widetilde{I}_t$, avoiding the need for fine-grained click--action video supervision.
\end{revision}

\begin{revision}
\paragraph{Anchor portfolio and state decision.}
The anchor states are not hard-coded actions or fixed scripts, but a replaceable and extensible visual portfolio $\mathcal{A}$ covering diverse poses, emotions, camera distances, and interaction contexts. Given the current visual state, the interaction history $H_{t-1}$, and the indexed anchor portfolio $\mathcal{A}$ with brief descriptions, the MLLM identifies the anchor that best matches the current state and the intended transition. Fine-grained dialogue, gestures, expressions, and object interactions remain handled by the action generator.
\end{revision}

\subsection{Video Model as a Digital Life Rendering Engine}

In this section, we elaborate on the architecture of our video generation pipeline, engineered to produce visual results that are immersive, coherent across multiple turns, and highly dynamic. To achieve these goals, our pipeline operates in three phases. Here, we detail the waiting video generation and action video generation phases.

Our simulation commences with the waiting video module, as we posit that a digital life should exhibit natural, subtle idle motions as a baseline state. To this end, we specialize a pre-trained video model by fine-tuning it on a dataset of natural and fluid waiting animations. This module is formulated as an Image-to-Video (I2V) generation task. We employ Low-Rank Adaptation (LoRA) \citep{hu2022lora} for this specialization, optimizing solely on an I2V loss. The use of LoRA obviates the need for extensive parameter retraining, allowing for lightweight and efficient integration.

Upon user-initiated interaction, the system transitions to the action video generation module. In this phase, the video model renders output based on semantic directives generated by the upstream MLLM. To create an immersive experience, the MLLM output is a structured representation containing two key components: a behavioral prompt (a textual description of the character's actions, expressions, and scene changes) and the dialogic response (the character's verbal response as text).

Consequently, the model for this stage must support synchronized audio-video generation, for which we leverage an existing open-source audio-text conditional video model \citep{kong2025let}. To satisfy this model's multimodal input requirements, the MLLM textual dialogic response is seamlessly synthesized into an audio waveform via an instant TTS engine, CosyVoice \citep{du2024cosyvoice}. This audio, jointly with the MLLM behavioral prompt, forms the complete conditional input used by the video model to generate the final audio-visual output.

\begin{revision}
\paragraph{Module switching and execution.}
The MLLM dynamically governs the switching logic. At each turn, Action Video first generates the speech-aligned response. AVC then conditions on the ending action segment and transitions to the selected target anchor. The waiting video associated with that anchor is played until the next interaction. When the selected anchor is the currently active one, AVC serves as a quality-restoring return; otherwise, it performs a macroscopic state transition. This explicit sequence makes state changes inspectable and controllable instead of allowing them to emerge implicitly from recent latent history.
\end{revision}

\subsection{Anchored Video Continuation}
\label{sec:avc}

Generating long-duration, multi-turn interaction videos of a digital life is hindered by two primary issues. The first challenge is quality degradation, as long-term, turn-by-turn generation leads to declining visual quality and character identity drift due to cumulative errors. The second challenge is limited motion dynamics, which occurs when the action model must simultaneously align video with audio (for lip-sync) and text (for actions). This strict synchronization requirement often forces a compromise in the range of motion, resulting in subdued movements that fail to capture larger, more dynamic actions like pose changes.

To address these issues, we propose an anchored video continuation mechanism. The core idea is to generate a dynamic transition from the end of each action segment to a pre-defined, high-quality anchor state. This design ensures long-term quality by resetting cumulative error, as the character periodically returns to a stable state. At the same time, by intentionally decoupling this transition from strict audio-alignment constraints, the mechanism allows for the generation of large, dynamic motions, significantly enhancing expressiveness.

A naive approach to achieve dynamic transition is to train a frame-based interpolation model to generate a transition video $v_{\text{trans}}$ by interpolating between the last frame of the action video $I_{\text{end}}$ and a preset anchor frame $I_{\text{anchor}}$. While this method connects the start and end points, we found its results to be suboptimal. Because the interpolation model only considers two static frames, it completely disregards the motion inertia of the character and the camera dynamics from the preceding video. This often results in a transition that feels disjointed and discontinuous from the previous action, as shown in our ablation study.

Therefore, we propose to train a segment-based continuation model. This model takes the final segment of the action video, which we denote as $v_{\text{context}}$, as a condition and continues it by generating a seamless transition to the anchor frame $I_{\text{anchor}}$. This generation process can be formulated as
\begin{equation}
v_{\text{trans}} \sim P(v\mid v_{\text{context}},I_{\text{anchor}}).
\label{eq:avc}
\end{equation}

We define $v_{\text{context}}$ as the last 25\% frames from the action video. Because the action videos themselves have varying lengths, the frame count of $v_{\text{context}}$ is also variable, which trains the model to handle different context lengths. We inject this condition in two ways: for detail preservation, all frames of $v_{\text{context}}$ and the anchor frame $I_{\text{anchor}}$ are compressed via VAE into a latent $Z_c\in\mathbb{R}^{c\times t\times h\times w}$ and concatenated with the noise latent $z_t$; simultaneously, for global context, we only extract the CLIP features from the first frame of $v_{\text{context}}$ and the anchor frame, injecting them into the DiT model via decoupled cross-attention.

\begin{revision}
\paragraph{Training-pair construction.}
We curate AVC training clips that contain meaningful changes in body pose, camera distance, or viewpoint rather than restricting the data to talking-head videos. The collection pipeline consists of multi-source collection, automated filtering, and annotation. We remove low-quality, poorly illuminated, or discontinuous clips using aesthetic and luminance filtering, retain single-shot videos with usable endpoints, and use Qwen2.5-VL to describe character appearance, action, scene content, and camera behavior. The resulting set contains 45k video--text pairs. The first quarter of each variable-length clip is used as the observed context and the final frame as the target anchor, producing diverse transition magnitudes without manually annotated transition labels. The waiting-video subset contains 3k clips with loop-compatible low-amplitude motion, such as breathing, blinking, minor head motion, hair movement, and environmental dynamics.

\paragraph{AVC implementation.}
AVC is initialized from the Wan 2.1 14B first--last-frame model and adapted with LoRA. Let $E_{\mathrm{vae}}$ denote its spatiotemporal VAE. The context frames and target anchor are encoded into $Z_c=E_{\mathrm{vae}}(v_{\text{context}}\cup I_{\text{anchor}})$ and concatenated channel-wise with the noisy video latent. A temporal mask $M\in\{0,1\}^{1\times T\times h\times w}$ marks every supplied context and anchor position:
\begin{equation}
M_j=\begin{cases}
1,&j\in\mathcal{T}_{\mathrm{context}}\cup\mathcal{T}_{\mathrm{anchor}},\\
0,&\text{otherwise}.
\end{cases}
\end{equation}
Unlike the original first--last-frame formulation, which preserves only the sequence boundaries, this mask supports an arbitrary conditioning segment. CLIP image features from the first context frame and the target anchor are injected through decoupled cross-attention. Using only the endpoint CLIP features limits memory growth, while the VAE latent retains frame-level appearance and motion over the full context. LoRA is applied to the attention projections and the original objective is retained.
\end{revision}

\section{Experiments}
\label{sec:experiments}

In this section, we demonstrate how DiVA addresses key limitations of prior models in supporting interactive digital life simulation via video models.

\begin{figure*}[!t]
    \centering
    \includegraphics[width=\textwidth]{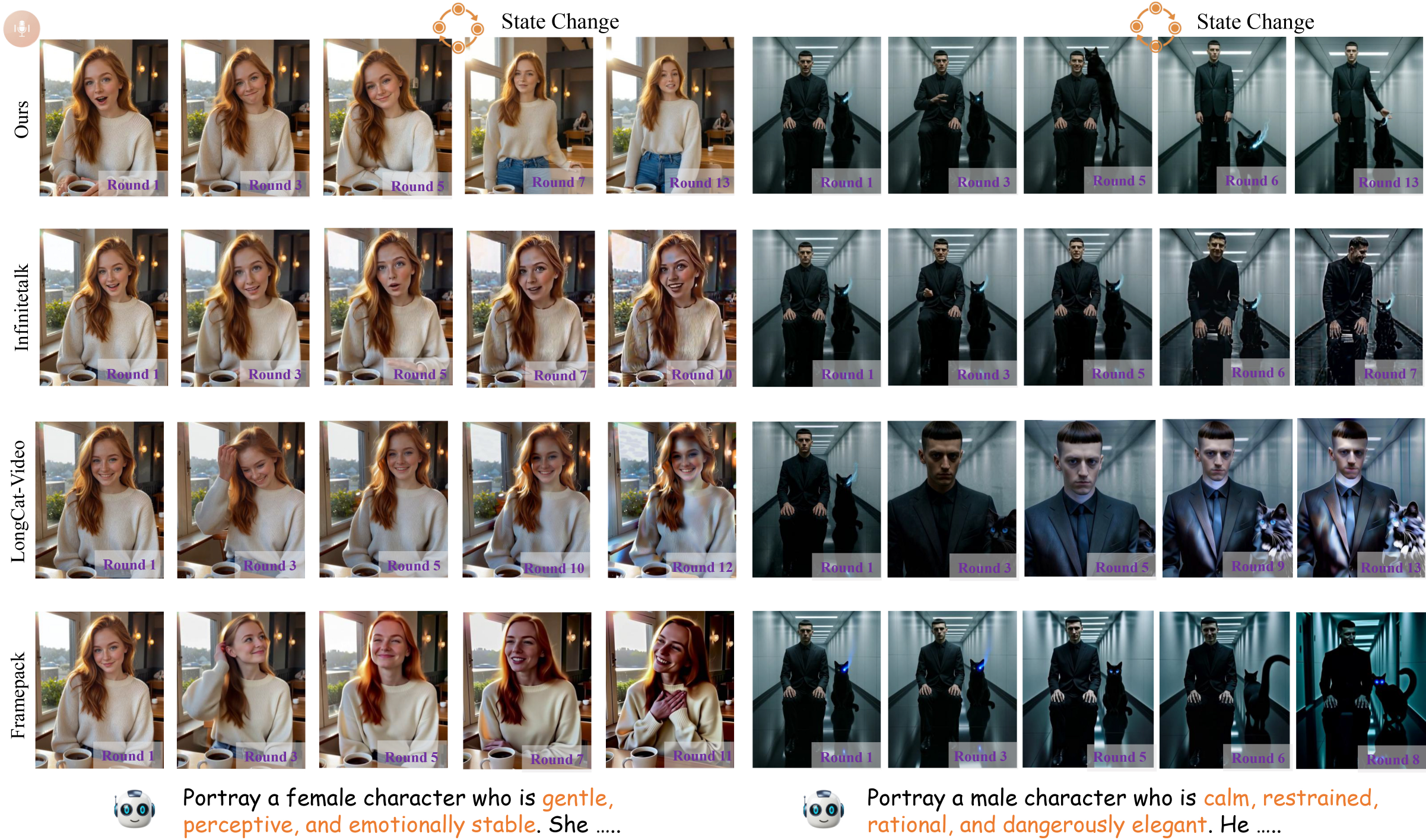}
    \caption{\textbf{Qualitative comparison on long-term digital life simulation.} Our method successfully generates high-quality, expressive, and drift-free results, accurately performing state transitions (e.g., the 3rd to 4th transition in both examples). In contrast, all baselines suffer from severe drift and fail to execute precise state transitions; for instance, InfiniteTalk's subject only partially stands up in the second example. Notably, all baselines exhibit severe drift in fewer than 10 interaction rounds (marked in the bottom-right of each image), whereas our method remains stable indefinitely. Best viewed zoomed in.}
    \label{fig:main_comparison}
\end{figure*}

\subsection{Implementation Details}
\label{sec:implementation}

We train our anchored video continuation model on our curated in-house dataset of 45k video--text pairs. All videos are processed at a resolution of $480\times832$. For training, the starting video is obtained by taking the first $1/4$ of the frames from a video, and the last frame of that video is extracted to serve as the interpolation target. Since the videos in our dataset vary in length, the starting videos used for training therefore have different lengths. This enables our model to support starting video inputs of varying lengths. During inference, we uniformly use 21 frames from the previous interaction video as the starting video.

Our base model is the Wan 2.1 14B FLF \citep{wan2025wan}. We train the model using LoRA with a rank of 128. The model is trained for 10k steps with a learning rate of $1\times10^{-5}$ and a linear warmup schedule. The entire training process is conducted on 48 NVIDIA H800 GPUs with a total batch size of 48. For the waiting video module, we also use Wan 2.1 14B as the base model, which allows our video base models to be shared across both tasks. We train this model using LoRA with a rank of 32. The model is trained for 5k steps on a dataset of 3k videos featuring subtle, live-wallpaper-style motions.

\begin{revision}
\paragraph{Evaluation protocol.}
Our long-horizon benchmark contains 60 multi-turn character simulations spanning diverse human, anime, stylized, and animal identities. Each sequence contains language interactions and explicit state transitions and comprises thousands of generated frames. All compared methods receive the same reference identity, prompts. FramePack and InfiniteTalk condition every turn after the first on the final frame of the previous turn. LongCat-Video follows its official continuation protocol, including its restoration and super-resolution stages. Methods without audio conditioning are evaluated for visual quality and state following rather than lip synchronization. Spatial grounding is evaluated on 50 interaction pairs with clicks on semantically distinct regions.
\end{revision}

\subsection{Comparison}
\label{sec:comparison}

We select the powerful base model FramePack \citep{zhang2025packing} (an I2V model with anti-drift designs to support long-video generation), InfiniteTalk \citep{yang2025infinitetalk} (a strong open-source audio-driven video generation model), and LongCat-Video \citep{team2025longcat} (a large pre-trained video continuation model) as our baselines, to evaluate our model's capabilities in maintaining persistent expressiveness in long-term interaction and accurate semantic understanding for state transition, respectively. We adopt the same prompts and a consistent turn-by-turn generation process. In each turn, we use the duration recommended by the corresponding method to avoid drift. For FramePack and InfiniteTalk, every turn after the first is conditioned on the final frame of the previous turn. For LongCat-Video, we follow its official guidelines, using the recommended video length from the previous turn for continuation, and its super-resolution and restoration functions are also applied in each turn. Since FramePack and LongCat-Video lack audio conditioning, we generate their results using only video prompts and focus the comparison on visual quality.

\subsubsection{Qualitative Results}

We present a comparison of our results against baselines in Fig.~\ref{fig:main_comparison}. These results were obtained from an interaction process spanning tens of rounds. To be concise, we visualize five of these rounds.

From the results, we can observe the following key characteristics. First, benefiting from our pipeline design, our method leverages a precise anchored video continuation module to accurately control the character's state during the interaction, thereby achieving natural state transitions. As illustrated in the figure, during the transition from the third interaction scene to the subsequent scenes, only our method successfully enables the character to assume a natural standing posture in both scenes. In contrast, FramePack attempts to stand but only achieves a half-standing state due to the uncontrollable nature of its generative model in the second example. In other examples, the baselines fail to follow the state transition entirely.

\begin{figure}[!t]
    \centering
    \includegraphics[width=0.95\linewidth]{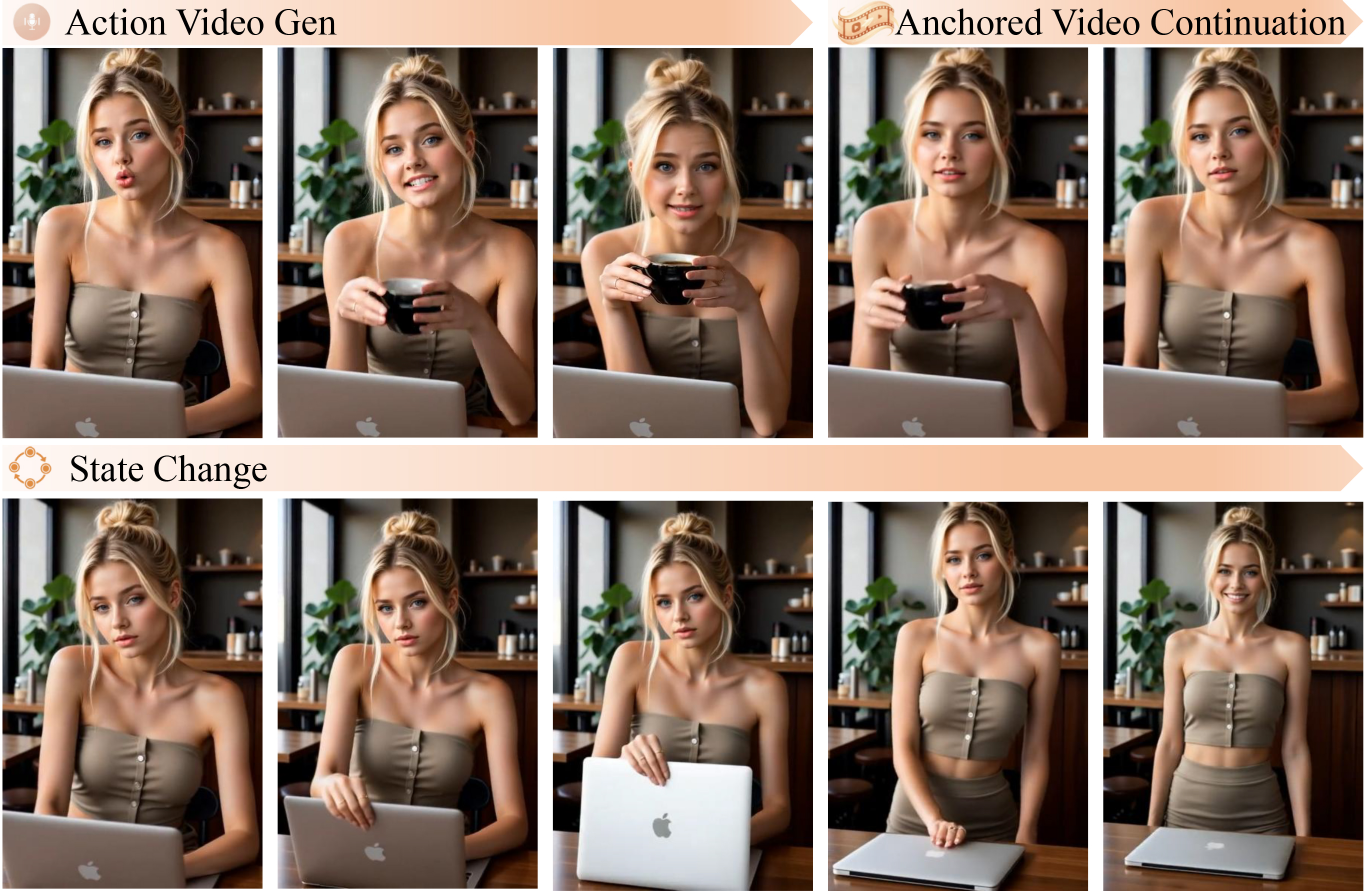}
    \caption{\textbf{Demonstration of our pipeline's high-quality continuation capability.} In this example, the model initially generates a coffee cup not present in the user-provided scene. Our AVC module seamlessly resolves this by generating a high-quality continuation of the character, plausibly setting the cup down. Subsequently, AVC executes a complex state transition: zooming out as the character stands and simultaneously closes the laptop, enabling more dynamic possibilities for character simulation.}
    \label{fig:consistency_capability}
\end{figure}

Second, our method consistently maintains high-fidelity visual quality and identity preservation throughout long-term interactions. When interacting with a character, it is often necessary to engage in ten or more rounds within a single scene. Our experiments reveal that while the baselines exhibit some anti-drift capabilities, visual degradation in the first few rounds may be subtle to humans; these minor degradations accumulate and are significantly amplified over multiple rounds. This leads to a substantial deterioration in color fidelity and character identity. We observe that the baseline models typically show noticeable degradation around the 5th round, followed by a sharp decline in quality near the 8th round.

These phenomena demonstrate that our video generation pipeline design significantly enhances model controllability. This not only facilitates more accurate state transitions but also ensures that the generation quality remains high and does not degrade, even over extended temporal durations.

To demonstrate the high-quality, consistent continuation capability of our pipeline, we present an example in Figure~\ref{fig:consistency_capability}. In this case, our model successfully handles interactions with newly generated objects (e.g., the character coherently picks up and later smoothly puts down a coffee cup via our AVC module). Furthermore, the model maintains continuity through pose changes. In the second row, for example, the character stands up with coherent camera movement. The model also logically integrates other actions, like closing the computer. This strong video continuation capability is essential for long-term, multi-turn interactions, proving our method's effectiveness in achieving high-quality, natural generation.

\subsubsection{Quantitative Results}

\begin{table*}[!t]
\centering
\caption{\textbf{Quantitative results.} The best result is shown in \textbf{bold}. Higher is better except for Quality Drift and Sync-D.}
\label{tab:compare}
\scriptsize
\setlength{\tabcolsep}{2.2pt}
\begin{tabular*}{\textwidth}{@{\extracolsep{\fill}}lccccccc@{}}
\toprule
\textbf{Method} & \shortstack{\textbf{Subject}\\\textbf{Consis.}$\uparrow$} & \shortstack{\textbf{Dynamic}\\\textbf{Degree}$\uparrow$} & \shortstack{\textbf{Image}\\\textbf{Quality}$\uparrow$} & \shortstack{\textbf{Quality}\\\textbf{Drift}$\downarrow$} & \textbf{Sync-C}$\uparrow$ & \textbf{Sync-D}$\downarrow$ & \shortstack{\textbf{Click}\\\textbf{Acc.}$\uparrow$} \\
\midrule
FramePack & 0.8979 & 0.4167 & 0.5961 & 0.1293 & -- & -- & -- \\
InfiniteTalk & 0.9202 & 0.3667 & 0.6143 & 0.0947 & 7.21 & 7.32 & -- \\
LongCat-Video & 0.8948 & 0.2833 & 0.6238 & 0.1195 & -- & -- & -- \\
\textbf{DiVA} & \textbf{0.9451} & \textbf{0.4833} & \textbf{0.6481} & \textbf{0.0341} & \textbf{7.53} & \textbf{7.16} & \textbf{0.8200} \\
\bottomrule
\end{tabular*}
\end{table*}

To evaluate our pipeline, we generated 60 videos across different scenes and assessed them for both general video quality and click response accuracy. For general quality, we adopted the VBench \citep{huang2024vbench} benchmark to measure subject consistency, dynamic degree, and image quality. Recognizing that significant quality differences can emerge between the beginning and end of videos, we also analyzed temporal drifting. Following FramePack \citep{zhang2025packing}, we report quality drift in Table~\ref{tab:compare}, which is defined as the absolute difference in image quality between the first 15\% and the last 15\% of the frames. We also report lip-sync metrics in the table. For click accuracy, we created 50 diverse interaction pairs using GPT with visually marked images and then manually evaluated whether the model's generated actions are logical and correct (e.g., ensuring a click on the ``right ear'' triggered a corresponding right-sided action).

The results indicate that the pre-trained LongCat-Video model maintains visual quality relatively well compared to other baselines, owing to its foundation in video continuation and the use of super-resolution. However, it tends to generate static scenes with minimal motion, resulting in a low dynamic degree. Although FramePack and InfiniteTalk can mitigate temporal drift to some degree, FramePack exhibits significant subject offsets over time, while its lack of audio-alignment constraints permits greater dynamism compared with other baselines. Additionally, all methods exhibit significant quality drift, while our pipeline achieves a strong balance among quality, motion, and lip-sync accuracy. Specifically, our AVC mechanism enables the model to revert to stable anchor states and dynamically transition between different anchors, which generates more expressive and superior dynamics, yielding strong all-around performance. For click accuracy, we found that using visual markers significantly improves the MLLM performance to a high precision. The MLLM consistently generates logical video prompts and responses, while occasional errors stem from the video model's fidelity in following these instructions. This straightforward approach validates our pipeline design, proving both its effectiveness and usability.

\begin{revision}
Quality Drift is computed as
\begin{equation}
\Delta M_{\mathrm{drift}}(V)=\left|M(V_{\mathrm{start}})-M(V_{\mathrm{end}})\right|,
\end{equation}
where $V_{\mathrm{start}}$ and $V_{\mathrm{end}}$ denote the first and last 15\% of the generated sequence. DiVA reduces Quality Drift from 0.0947 for InfiniteTalk, the strongest baseline on this metric, to 0.0341, a relative reduction of approximately 64\%. Importantly, this improvement is not obtained by suppressing motion: DiVA also achieves the highest Dynamic Degree of 0.4833.
\end{revision}

\begin{revision}
\subsection{Additional Audio-Driven Avatar Baselines}

We further evaluate strong audio-driven avatar  StableAvatar and WanS2V under the same multi-turn DLS protocol. Table~\ref{tab:avatar_baselines} shows that both models accumulate more quality drift than InfiniteTalk and remain substantially behind DiVA. These methods are valuable when the primary requirement is uninterrupted speech-driven avatar rendering with limited state change. DiVA targets a different operating point in which speech-aligned responses are interleaved with explicit large-motion transitions and controlled visual resets.
\end{revision}

\begin{table}[!t]
\centering
\caption{\rev{Quality Drift for additional audio-driven avatar baselines. Lower is better.}}
\label{tab:avatar_baselines}
{%\color{red}
\begin{tabular}{lc}
\toprule
\textbf{Method} & \textbf{Quality Drift}$\downarrow$ \\
\midrule
StableAvatar \citep{tu2025stableavatar} & 0.1393 \\
WanS2V \citep{gao2025wan} & 0.1151 \\
InfiniteTalk \citep{yang2025infinitetalk} & 0.0947 \\
\textbf{DiVA} & \textbf{0.0341} \\
\bottomrule
\end{tabular}}
\end{table}

\begin{revision}
\subsection{The Effectiveness of the Decoupled Pipeline}

As quantitatively demonstrated in Table~\ref{tab:compare}, our decoupled architecture effectively addresses the stability--expressiveness trade-off inherent in long-term video generation. General long-video baselines struggle with identity preservation over time (reflected in higher Quality Drift), as their generated states are repeatedly propagated without a resetting mechanism. Conversely, while purely audio-driven avatar models achieve better local identity preservation and speech alignment, their strong audio coupling severely restricts macroscopic pose and camera changes, leading to suboptimal Dynamic Degree scores.

Furthermore, this decoupled design explicitly exposes the state-switching logic, contributing directly to the system's robust long-term metrics. Instead of allowing state transitions to emerge implicitly from a continuous stream of recent generated frames---which our experiments show exacerbates error accumulation---the explicit sequence of MLLM anchor selection, Action Video generation, and AVC transition ensures the sequence periodically resets to high-quality visual states. This inspectable mechanism is the primary driver behind the reduced long-term degradation observed in our quantitative evaluations.
\end{revision}

\subsection{Ablation Study}
\label{sec:ablation}

\begin{figure*}[!t]
    \centering
    \includegraphics[width=\textwidth]{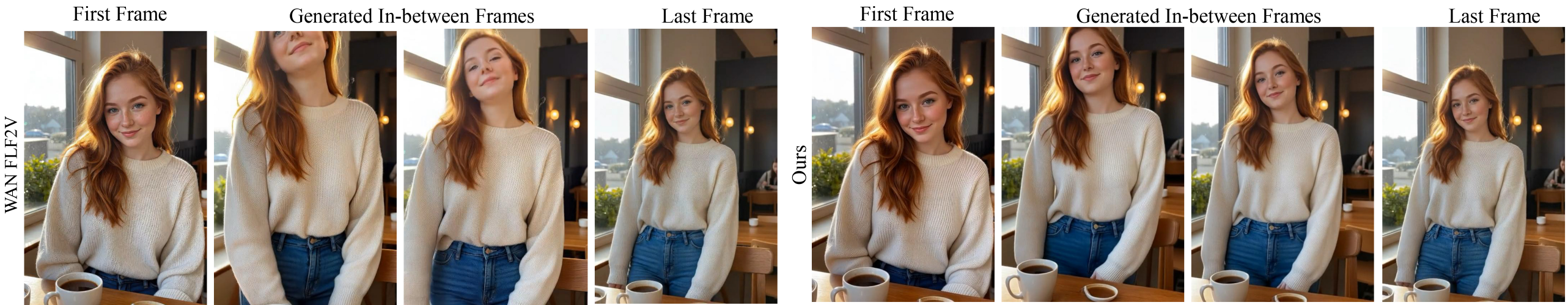}
    \caption{\textbf{Comparison of video frame interpolation methods.} We compare our method against Wan FLF2V. Our approach generates more natural camera movements and character pose transitions, while the base Wan method (relying solely on the last frame) exhibits less harmonious camera work and character motion. Our method effectively avoids these inconsistencies. We recommend viewing the video samples for a clear demonstration of camera and character motion.}
    \label{fig:ablation_interpolation}
\end{figure*}

A key component of our design is the anchored video continuation module, which generates the intermediate frames between a previously generated video clip and anchor frames. To validate AVC, we conduct a comparative analysis against the strong Wan FLF2V baseline, with qualitative results presented in Figure~\ref{fig:ablation_interpolation}. We observe that conventional single-frame interpolation models lack prior knowledge of the character motion and camera dynamics established in the preceding video. Consequently, their interpolated results often suffer from unnatural transitions and camera movements. For instance, in the second example, the character generated by Wan FLF2V suddenly moves out of the frame and executes a rapid, unnatural shift between two distinct actions. In contrast, our model produces a significantly more natural and coherent transition. These temporal artifacts are difficult to capture with static images or
general frame-level metrics and are substantially more pronounced in motion, 
we therefore provide the complete comparisons in the supplementary
video.

\begin{table}[!t]
\centering
\caption{\textbf{User study results on transition naturalness.} Preference scores for Transition Fidelity (TF) and Physical Plausibility (PhyP). The best results are in \textbf{bold}.}
\label{tab:user_study}
\scriptsize
\setlength{\tabcolsep}{4pt}
\begin{tabular}{lcc}
\toprule
\textbf{Method} & \shortstack{\textbf{Transition}\\\textbf{Fidelity}$\uparrow$} & \shortstack{\textbf{Physical}\\\textbf{Plausibility}$\uparrow$} \\
\midrule
Wan FLF2V & 0.233 & 0.283 \\
\textbf{DiVA} & \textbf{0.767} & \textbf{0.717} \\
\bottomrule
\end{tabular}
\end{table}

\begin{revision}
\paragraph{Human evaluation of transition naturalness.}
We further conducted a user study since complex visual dynamics such as camera trajectory smoothness and physical motion continuity are not reliably captured by frame-level metrics. Thirty participants evaluated 60 transition videos spanning 10 character styles and species. Each participant compared DiVA with Wan FLF2V in terms of Transition Fidelity, which assesses camera stability and framing continuity, and Physical Plausibility, which assesses the realism of character and camera motion. As shown in Table~\ref{tab:user_study}, DiVA is preferred in 76.7\% of comparisons for Transition Fidelity and 71.7\% for Physical Plausibility.
\end{revision}

\begin{figure*}[!t]
    \centering
    \includegraphics[width=\textwidth]{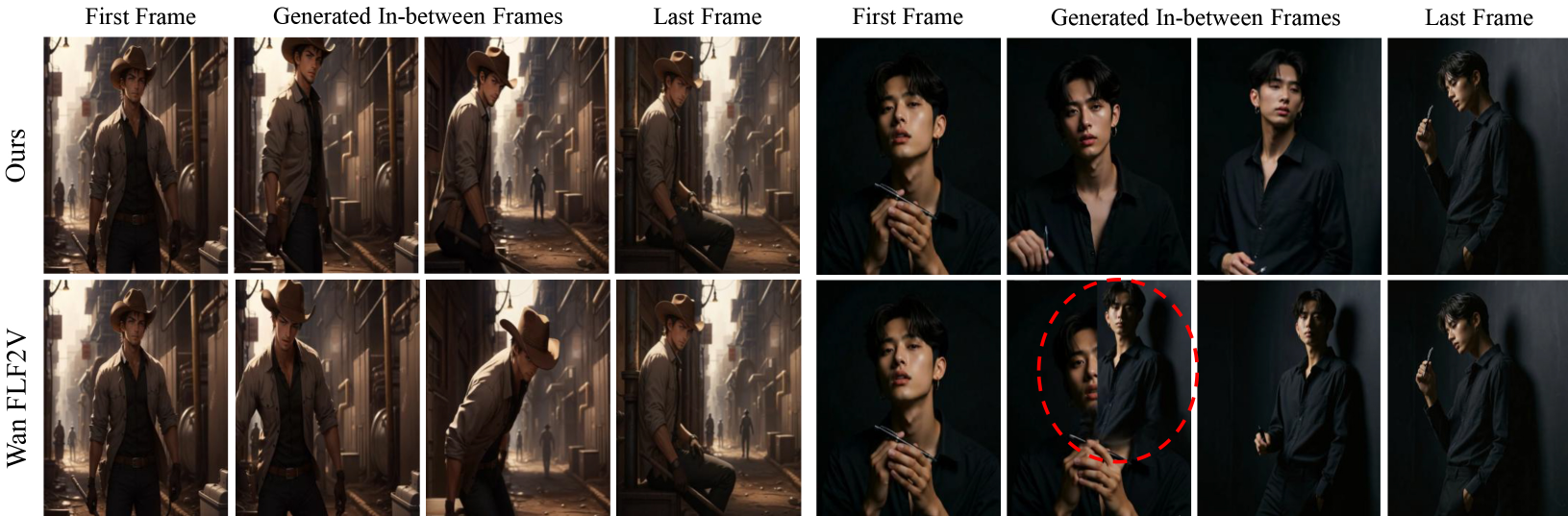}
    \caption{\rev{Qualitative gallery of transition naturalness across diverse character styles. Under substantial pose and camera shifts, Wan FLF2V often introduces an abrupt shot change, whereas DiVA maintains continuous motion and camera flow.}}
    \label{fig:transition_gallery}
\end{figure*}

\begin{revision}
Figure~\ref{fig:transition_gallery} provides a broader qualitative gallery covering photorealistic humans, stylized characters, and anime. The examples further demonstrate that segment conditioning remains effective across appearance domains and substantial viewpoint changes.
\end{revision}

\noindent\textbf{Alternative Designs for CLIP Injection.}
In addition to latent concatenation, we leverage the image encoder of CLIP \citep{radford2021learning} to extract high-level feature representations from the condition images. These representations are subsequently injected into the DiT model via decoupled cross-attention. Let the extracted visual patch features be denoted as $\mathbf{f}_c\in\mathbb{R}^{l\times n\times c}$, where $l$ represents the number of condition images, $n$ is the number of patches per image, and $c$ is the feature dimension. Directly concatenating the features of all condition images along the sequence dimension would result in an excessively long context sequence. Furthermore, adjacent frames in a video segment typically share highly redundant semantic information. To address this, we explore two strategies for processing the CLIP embeddings: (1) \emph{Keyframe Selection}, where we retain only the first frame of the video segment and the stable-state anchor frame; and (2) \emph{Temporal Averaging}, where we compute the mean of the CLIP features across all frames in the video segment to obtain a condensed representation in $\mathbb{R}^{1\times n\times c}$, which is then concatenated with the feature of the stable-state anchor frame.

A comparison of general quality across 100 generated transition videos reveals negligible performance differences between the two methods. This implies that the global CLIP condition is robust to frame selection (multi-frame versus first-frame), as sufficient information is integrated by the latent condition. We therefore use keyframe selection for efficiency.

\begin{table}[!t]
\centering
\caption{\rev{Anchor configuration ablation. Historical anchors are generated frames from the preceding interaction history rather than curated visual states.}}
\label{tab:anchor}
\scriptsize
\setlength{\tabcolsep}{1.8pt}
{%\color{red}
\begin{tabular*}{\linewidth}{@{\extracolsep{\fill}}lccc@{}}
\toprule
\textbf{Configuration} & \shortstack{\textbf{Subject}\\\textbf{Consis.}$\uparrow$} & \shortstack{\textbf{Dynamic}\\\textbf{Degree}$\uparrow$} & \shortstack{\textbf{Quality}\\\textbf{Drift}$\downarrow$} \\
\midrule
1 Anchor (static) & 0.8842 & 0.4145 & 0.0911 \\
Historical anchors & 0.8633 & 0.4512 & 0.1425 \\
\textbf{5 Curated} & \textbf{0.9451} & \textbf{0.4833} & \textbf{0.0341} \\
\bottomrule
\end{tabular*}}
\end{table}

\begin{revision}
\subsection{Anchor-State Analysis}

We compare three anchor configurations in Table~\ref{tab:anchor}. A single static anchor provides a reliable reset but forces diverse interactions back to one pose, reducing dynamic range and causing unnatural spatial warping. Historical anchors use recent generated frames, similar to a single-stream continuation strategy; they preserve more recent motion but also preserve and amplify generated errors. Five curated anchors provide both pose coverage and clean reset points, producing the strongest subject consistency, highest Dynamic Degree, and lowest Quality Drift.
\end{revision}

\begin{figure}[!t]
    \centering
    \includegraphics[width=0.95\linewidth]{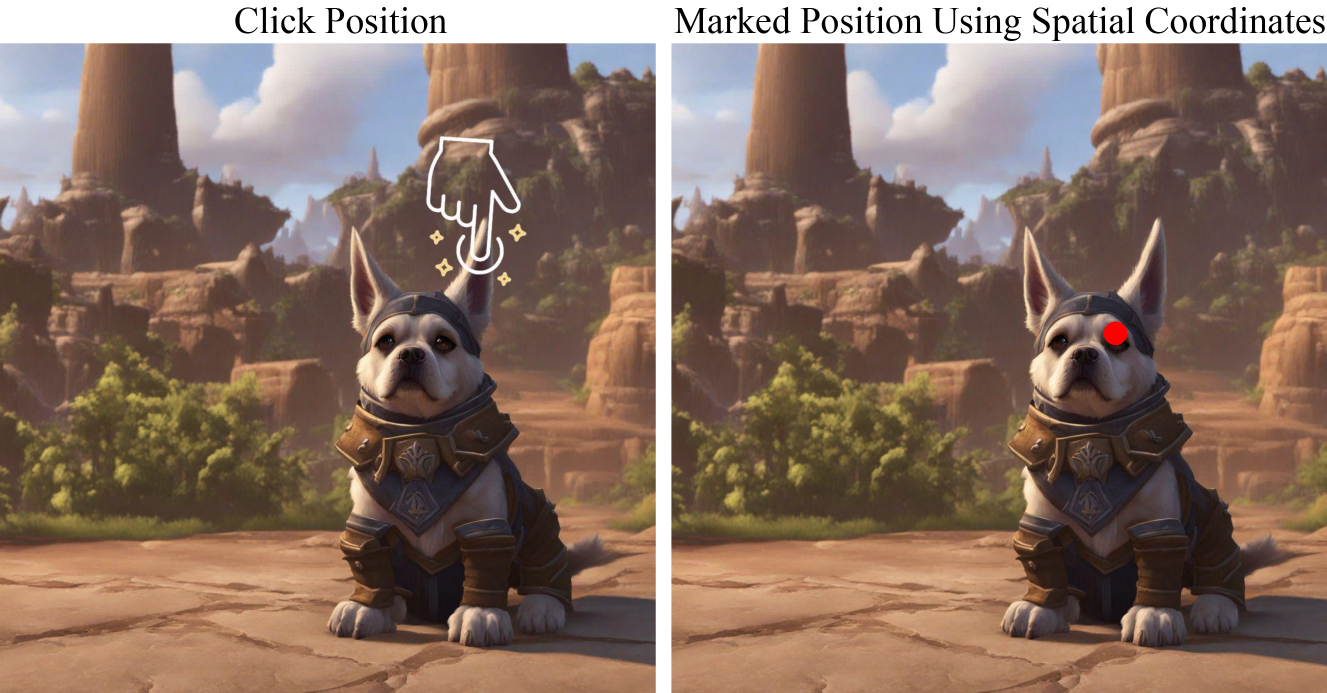}
    \caption{\rev{Ablation of spatial click input. Direct coordinate input struggles with precise localization, leading to semantic confusion between adjacent features (e.g., ear versus eye). Visual markers provide explicit spatial grounding for the MLLM.}}
    \label{fig:click_grounding}
\end{figure}

\begin{figure*}[!t]
    \centering
    \includegraphics[width=0.85\textwidth]{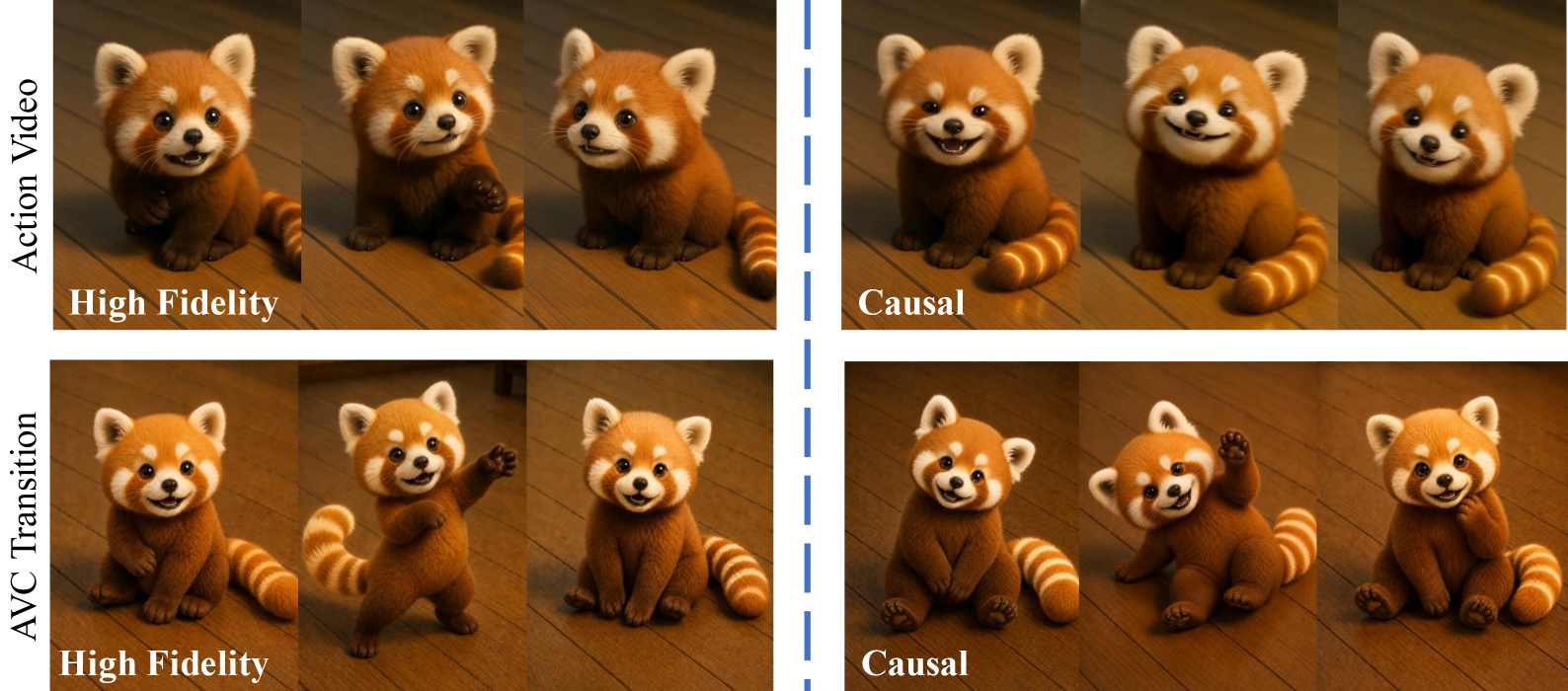}
    \caption{\rev{Comparison between high fidelity configuration and causal distillation.}}
    \label{fig:latency}
\end{figure*}

\begin{revision}
These results show that anchor quality and state coverage are both important. A reset state must be sufficiently clean to remove accumulated error, but the portfolio must also cover enough macroscopic poses to avoid forcing every interaction through one rigid endpoint. The five-anchor configuration provides the best balance between expressive state changes and long-term stability.
\end{revision}

\begin{revision}
\subsection{Spatial Click Grounding}

Figure~\ref{fig:click_grounding} compares raw coordinate prompting with our visual-marker strategy. Numerical coordinates require the MLLM to learn an explicit coordinate-to-pixel correspondence and frequently confuse nearby semantic regions. Rendering the click directly on the current frame turns the problem into visual referring and raises click-response accuracy to 0.82 on 50 interaction pairs. The mechanism requires no additional video-model training and can be applied to arbitrary character categories.
\end{revision}

\noindent\textbf{Inference Cost and Speed.}
Our framework is highly adaptable, offering a flexible trade-off between inference speed and generation quality. By integrating INT8 quantization, feature caching, and step distillation, our primary high-fidelity configuration generates high-quality and expressive motion in four sampling steps, requiring approximately 4\,s per step on an NVIDIA H800 GPU. Our architecture also supports causal distillation for real-time inference. Since real-time causal generation often trades fine-grained visual quality, lip-sync accuracy, and motion magnitude for higher throughput, following prior work~\cite{cheng2025animegamer}, we use the high-fidelity configuration as our primary setting to provide a more immersive simulation experience, while retaining a causal configuration for latency-sensitive deployment.

\begin{figure*}[!t]
    \centering
    \includegraphics[width=\textwidth]{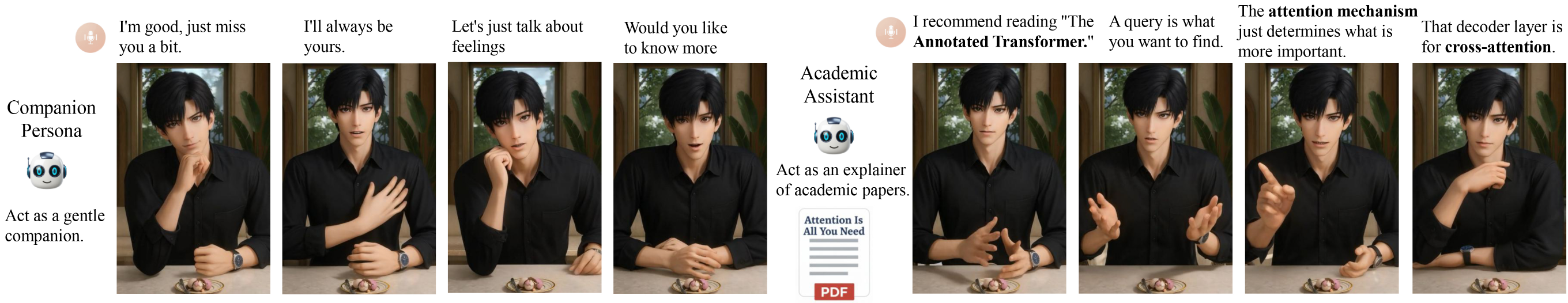}
    \caption{\textbf{Persona-driven interaction.} Our method dynamically defines distinct character personas, enabling varied roles with appropriate verbal and non-verbal responses. The examples include a companion persona offering empathetic interaction with fitting gestures and an academic assistant explaining complex PDFs (e.g., ``Attention Is All You Need''). The assistant provides factually correct explanations and uses gestures that enhance clarity.}
    \label{fig:applications}
\end{figure*}

\begin{figure*}[!t]
    \centering
    \includegraphics[width=\textwidth]{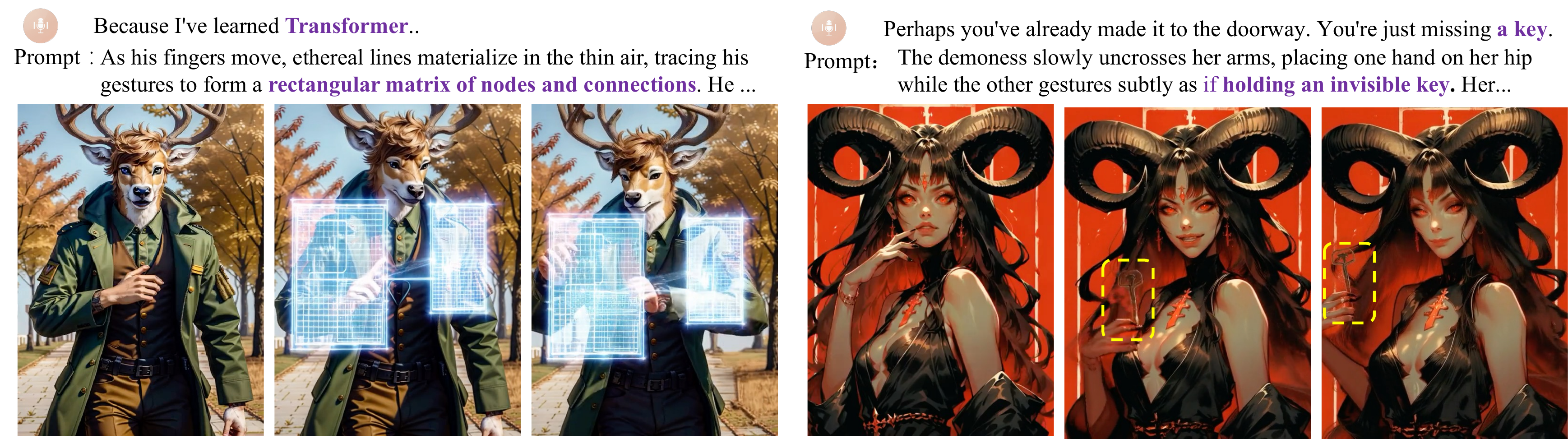}
    \caption{\textbf{Discussion of visual hallucination.} The video model's powerful generative imagination allows it to follow complex prompts, such as visualizing ``Transformer'' as a square matrix or hallucinating a key. These cases demonstrate advanced prompt-following fidelity, even when the results defy physical common sense.}
    \label{fig:visual_hallucination}
\end{figure*}

\begin{revision}
The overall expressiveness of DiVA is not determined by the action stage alone. Our decoupled AVC module is free from strict audio-alignment constraints and can therefore perform large-scale pose and camera transitions with high visual fidelity. As shown in Figure~\ref{fig:latency}, this design preserves macroscopic expressiveness even when the causal action video exhibits restrained motion. Under the real-time configuration, evaluated on NVIDIA H800 GPUs, DiVA achieves an end-to-end response latency of approximately 1.5\,s and an interactive rendering throughput of approximately 15 FPS. Under the same evaluation protocol, representative baselines such as FramePack and InfiniteTalk require approximately 28.5\,s and 24.5\,s per interaction turn, respectively. X-Streamer reports a higher throughput of 25 FPS, but primarily targets video-call-style portrait streaming at a resolution of $256\times256$, rather than the large pose and camera transitions supported by DiVA.
\end{revision}

\subsection{Applications of DiVA}

Our model inherently understands multimodal knowledge and can simulate diverse behavioral styles, enabling it to adeptly perform a wide range of tasks. In Figure~\ref{fig:applications}, we demonstrate several such scenarios. In the first example, the user defines the digital character as a companion. In this role, the character's actions and linguistic responses adopt a gentle and supportive demeanor, with each reply consistently aligning with the defined companion persona. In the second example, the character is defined as an academic assistant and provided with external documentation to lecture on. In this context, the character adopts a more formal posture, and its responses are factually accurate; for instance, it correctly recommends the Annotated Transformer \citep{rush2018annotated} and provides a precise explanation of the attention mechanism. This versatility in role-playing highlights the significant potential of our model for future digital character applications.

\subsection{Discussion of Visual Hallucination}

While video models achieve high realism in character simulation, they are prone to illogical artifacts akin to clipping in 3D modeling, which we term visual hallucination. As illustrated in Figure~\ref{fig:visual_hallucination}, we present two instances. In the first case, the model faithfully renders a rectangular matrix with nodes based on a transformer-related prompt, yet the resulting object lacks physical grounding. In the second case, the character interacts with a key that suddenly appears, although it is absent from the original frame. These instances demonstrate the model's robust instruction-following. This highlights a trade-off where the model's creative and imaginative response, often valued by users, takes precedence over strict physical realism. Balancing this trade-off remains a direction for future research.

\begin{revision}

\subsection{Limitations and Boundary Cases}

DiVA inherits fine-grained generation failures from the underlying video models, especially for physically unusual interactions. When the selected target anchor differs extremely from the current action segment, AVC generally preserves a smooth transition but may become cross-fade-like rather than producing fully plausible intermediate motion. Future work may combine, stronger physical priors, and online state discovery to enlarge the reachable visual state space while retaining explicit quality control.
\end{revision}

\FloatBarrier

\section{Conclusion}
\label{sec:conclusion}

We introduce DiVA, a novel deep simulator designed for building an open-ended interactive digital character world. This represents a crucial step in pushing video generation beyond passive clips toward long-term, immersive digital life simulation. DiVA leverages an MLLM as a router to dynamically process complex user linguistic and spatial input, enabling a highly persona-driven, open-ended simulation experience. To maintain high dynamics and visual stability during continuous interaction, we designed a three-part coupled video pipeline (consisting of waiting, action, and anchored video continuation). This pipeline, featuring the proposed AVC mechanism, effectively suppresses temporal identity drift while simultaneously unlocking greater dynamic expression. Ultimately, DiVA provides a robust foundation for simulating long-duration, controllable characters, advancing the future of generative interactive worlds.

\section*{Declarations}

\subsection*{Competing Interests}
The authors declare that they have no relevant financial or non-financial interests to disclose.

\subsection*{Ethics Approval and Consent to Participate}
The user study collected only anonymous preference judgments on generated videos and did not collect identifiable or sensitive personal information. Participation was voluntary, and informed consent was obtained from all participants prior to the study. The study was conducted in accordance with the applicable institutional requirements.

\subsection*{Data Availability}
The public datasets and pretrained models used in this study are cited in the manuscript. Evaluation materials and generated results will be released with the project materials upon publication.

\subsection*{Code Availability}
The inference and evaluation code, model adaptations, and prompt templates will be made publicly available with the project materials upon publication.

\subsection*{Author Contributions}
Cheng Chen led the conceptualization and technical development of the method, implemented the system, conducted the experiments, analyzed and visualized the results, and prepared the original manuscript. Hao Ouyang also contributed substantially to methodology development, result interpretation, and manuscript revision. Qiuyu Wang, Ka Leong Cheng, and Wen Wang contributed to the methodology, system development, and experimental evaluation. Yihao Meng, Hanlin Wang, Yixuan Li, Jiacheng Wei, and Zhenshan Tan contributed to data preparation, implementation, evaluation, and result analysis. Yanhong Zeng and Yujun Shen provided technical guidance, computational resources, and manuscript revision. Guosheng Lin and Fayao Liu supervised the research and contributed to the conceptual development and revision of the manuscript. All authors discussed the results, reviewed the manuscript, and approved the final version.

\bibliography{references}

\end{document}